\documentclass[journal,twoside,web]{ieeecolor}
\usepackage{generic}
\usepackage{cite}
\usepackage{amsmath,amssymb,amsfonts}
\usepackage{algorithmic}
\usepackage{graphicx}
\usepackage{textcomp}
\usepackage{comment}
\usepackage{makecell}
\usepackage{multirow}
\usepackage{comment}
\usepackage{caption}
\usepackage{orcidlink}

\newcolumntype{P}[1]{>{\centering\arraybackslash}p{#1}}
\def\BibTeX{{\rm B\kern-.05em{\sc i\kern-.025em b}\kern-.08em
    T\kern-.1667em\lower.7ex\hbox{E}\kern-.125emX}}
{Author \MakeLowercase{\textit{et al.}}: SDFA: Structure Aware Discriminative Feature Aggregation for Efficient Human Fall Detection in Video}
\begin{document}
\title{SDFA: Structure Aware Discriminative Feature Aggregation for Efficient Human Fall Detection in Video}

\author{Sania Zahan \orcidlink{0000-0002-5603-2170}, Ghulam Mubashar Hassan \orcidlink{0000-0002-6636-8807}, \IEEEmembership{Member, IEEE}, and Ajmal Mian \orcidlink{0000-0002-5206-3842}, \IEEEmembership{Senior Member, IEEE}
%\thanks{Manuscript received 2 July 2022; revised 5 October 2022; accepted 26 October 2022. This work was supported by the Australian Research Council, grant number DP190102443. The work of Sania Zahan was supported in part by a University Postgraduate Award and in part by the University of Western Australia International Fee Scholarship. The work of Ajmal Mian was supported by the Australian Research Council Future Fellowship Award under Project FT210100268 funded by the Australian Government. Paper no. TII-22-2925.(Corresponding author: Sania Zahan.)}
%\thanks{The authors are with the Department of Computer Science and Software Engineering, The University of Western Australia, Crawley WA 6009, Australia (e-mail: sania.zahan@research.uwa.edu.au (corresponding author); ghulam.hassan@uwa.edu.au; ajmal.mian@uwa.edu.au).}
%\thanks{Color versions of one or more figures in this article are available at https://doi.org/10.1109/TII.2022.3221208.}
%\thanks{Digital Object Identifier 10.1109/TII.2022.3221208.}
}

\maketitle

\begin{abstract}
Older people are susceptible to fall due to instability in posture and deteriorating health. Immediate access to medical support can greatly reduce repercussions. Hence, there is an increasing interest in automated fall detection, often incorporated into a smart healthcare system to provide better monitoring. Existing systems focus on wearable devices which are inconvenient or video monitoring which has privacy concerns. Moreover, these systems provide a limited perspective of their generalization ability as they are tested on datasets containing few activities that have wide disparity in the action space and are easy to differentiate. Complex daily life scenarios pose much greater challenges with activities that overlap in action spaces due to similar posture or motion. To overcome these limitations, we propose a fall detection model, coined SDFA, based on human skeletons extracted from low-resolution videos. The use of skeleton data ensures privacy and low-resolution videos ensures low hardware and computational cost. 
Our model captures discriminative structural displacements and motion trends using unified joint and motion features projected onto a shared high dimensional space. 
Particularly, the use of separable convolution combined with a powerful GCN architecture provides improved performance. 
Extensive experiments on five large-scale datasets with a wide range of evaluation settings show that our model achieves competitive performance with extremely low computational complexity and runs faster than existing models.
\end{abstract}


\section{Introduction}
\label{sec:introduction}
\IEEEPARstart{F}{alling} is inevitable at any stage of life. However, deteriorating physical and mental health makes older people more susceptible to fall and the consequences can be tragic, especially for those who live alone. Delays in treatment can cause severe long-term difficulties. Injuries such as fractures, open wounds, intracranial injuries etc, are quite common as a result of a severe fall and can cause chronic disabilities. Besides, falling often leads to fear of fall or Psychomotor Disadaptation Syndrome (PDS) affecting the quality of life by reducing confidence in mobility \cite{MATHON2017e50}.

Fall is often the first sign of a new or worsening health condition such as vision or cognitive impairments, heart disease, Osteoarthritis, Parkinson’s disease, etc \cite{disease_2021}. Fall is the second major cause of accidental death worldwide after road traffic injuries \cite{WHO_FALL}. According to a UN report, the global aging population will double by 2050, exceeding 2 billion \cite{UN_age}. It will create significant pressure on healthcare infrastructure and budget, especially in low-income countries. Besides, medical cost estimations for fall-related injuries often consider immediate support only and do not reflect long-term consequences. Therefore, proper monitoring in care facilities and home is critical to detect falls as well as to provide valuable insights into health conditions.

Conventional fall detection systems are based on wearable devices \cite{Alarifi_2021, Carlier2020FallDA, Liu_ICAIBD} that use sensors to measure deviation in structural displacements. If the measured deviation exceeds a predefined threshold, it is classified as fall. However, people with different physiques and lifestyles have varied movement characteristics. Therefore,  constrained thresholding generates high rate of false alarms for some and none for others. In addition, daily activities that involve fast movements with higher fluctuation in motion, such as shaking an object or dusting, can also cause false alarms \cite{Medical_alert_2021}. Besides, it can be inconvenient  as well as an extra burden for older people as they become more forgetful. The risk is even greater for people with dementia who are more prone to fall \cite{dimentia_fall}.

To overcome the limitations of wearable fall detection devices, some methods use video-based monitoring by installing a conventional camera that has a wide field of view \cite{ASLAN20151023, bian2015, Chen2020, Kottari2020}. However, direct use of video feeds violates individual privacy and creates legal issues. Therefore, some studies applied encoding techniques to obscure intelligibility of the video. However, this does not ensure privacy as reverse engineering to retrieve the identity is still possible. 

Recently, skeleton-based human action analysis has gained heightened popularity with the introduction of Kinect systems \cite{liu2020disentangling, Keskes_2021} and pose estimation algorithms \cite{Wang2020}. Skeleton data does not contain appearance information, and is robust to variations in viewpoint, lighting conditions, backgrounds and other noises. Moreover, reverse engineering of skeleton data to retrieve the original video is not possible, making it the most suitable modality for monitoring actions without compromising privacy. Skeleton joints have lower dimensions, compared to video, and yet convey rich motion information and have a consistent structural pattern over time. Therefore, it enables effective extraction of complex interactions without heavy computational requirements. 

Skeletons can be easily captured using Kinect or various pose estimation algorithms such as OpenPose. Though Kinect produces 3D skeletons with more detailed information of orientation and scale, it is relatively expensive. To benefit from low cost cameras, pose estimation algorithms are increasingly exploited in recent works \cite{XU2020123205, Wang2020} which can directly extract 2D skeletons from video frames ensuring both privacy and cost-effectiveness. However, this domain has not been explored for fall detection. Besides, existing skeleton based fall detection methods utilize standard CNN networks that use generalized filters for feature aggregation \cite{Tsai_2019, Wang2020, XU2020123205}. Considering the natural resemblance of human skeletons to graphs, graph convolutional networks (GCN) are more intuitive and can provide structure-specific attention. Therefore, in this paper, we propose a fall detection model based on OpenPose \cite{openpose2019} to extract 2D skeleton sequences from RGB videos and GCN to encode robust spatio-temporal features. We evaluated our model on large scale daily life activities including fall. Our main contributions are summarized as follows: 

\begin{itemize}
  \item \textbf{Cost effectiveness:} We propose an efficient fall detection system that uses 2D skeletons extracted from common videos, facilitating the use of low specification cameras for monitoring. We explored different resolutions to verify the robustness of our system against video quality. Experimental results corroborate system efficiency across multiple datasets with different frame resolutions and evaluation settings. To our knowledge, this is the first work to evaluate cross fall along with other evaluations where the performance in detecting unseen fall types is analyzed.
  
  \item \textbf{Memory efficiency:} We propose adaptive graph-based feature aggregation with two consecutive separable convolutional neural networks to capture slow and fast temporal articulations. This allows our system to extract powerful features without employing exorbitantly deep or wide network layers. Therefore, floating point operations per second (FLOPS) and the number parameters of our model are extremely low as compared to other existing state-of-the-art methods and without any noticeable drop in performance.
  
  \item \textbf{Generalization:} Our proposed architecture contains random spatio-temporal masking that forces our model to focus on discriminative motion cues by randomly dropping joints and frames. Thus, the model learns to adaptively refocus spatio-temporal attention to achieve better-generalized feature expressions.
\end{itemize}

The remaining paper is organized as follows: Section~\ref{litreature_rev} explores existing fall detection methods. Section~\ref{Methodology} gives details of the proposed architecture. Section~\ref{Experiments} represents experimental results, comparisons with existing methods, and discusses the significance of model segments through ablation studies. Section~\ref{Discussion} provides discussion and conclusions are given in Section~\ref{Conclusions}.

%%%%%%%%%%%%%%%%%%%%%%%%%%%%%%%%%%%%%%%%%%
\section{Related Works}\label{litreature_rev}
Most of the existing literature exploits handcrafted features calculated either using signals from wearable devices or video frames from vision based systems. Though, they report high accuracy, their reliability is limited for practical applications as there is little evidence of their performance in an everyday household with a dynamic and uncertain environment.

Wearable devices utilize different sensors such as inclinometers, gyroscopes, accelerometers etc, which provide information about vibration, displacement and angular velocity. These factors are most prominent for \emph{Fall} but also common in other activities of daily life. To improve overall performance, Nahian et al. \cite{Nahian_access} used multivariate sensor signals exploiting multiple information sources. Abdulaziz et al. used feature selection algorithms to extract prominent features \cite{Alarifi_2021}. However, these systems solely rely on handcrafted features and are susceptible to noise.

For vision-based systems, researchers mainly focused on videos. Most of these systems approached the task in two segments: feature extraction and then fall detection from the extracted features. Muzaffer et al. proposed a shape-based characterization to calculate geometric features of depth videos \cite{ASLAN20151023}. Bian et al. applied silhouette extraction to track body parts over consecutive frames to analyze head joint trajectory for fall motion detection using SVM \cite{bian2015}. However, this trajectory is similar for several other activities such as sitting, lying down, etc, which creates overlaps in patterns. Therefore, traditional classifiers such as SVM may perform reasonably well for fewer activities but with the increase in activity types, the performance will decrease. Though, the use of complex non-linear decision boundaries may increase accuracy to a certain extent, it will inadvertently introduce overfitting problem. 

Deep learning methods are more efficient in learning subtle structures and interactions. Carlier et al. used dense flow and global/local feature maps from videos to train a fully connected network \cite{Carlier2020FallDA}. Similarly, most deep learning methods used different convolutional networks to extract feature vectors from videos \cite{Chen2020, Kottari2020}. Despite superior performance, directly utilizing video feed is not acceptable for monitoring daily life activities. Privacy concern restricts widespread deployment of such systems, leading some researchers to explore compression or facial region anonymization to reduce visual perception \cite{Liu_ICAIBD, MA201944}. 

Contrarily, skeleton-based fall detection models provide a more practical way to estimate structure and motion trends without compromising privacy. It alleviates the direct use of video frames. Particularly, skeletons do not provide any person-specific details. Furthermore, GCN architectures are adept in adjusting weight distribution to extract subtle joint dependency from low dimensional skeleton data outperforming existing video-based systems \cite{sania2021}. However, the GCN based model provided in \cite{sania2021} uses 3D skeletons captured from Kinect sensor which limits its usability. In this paper, we propose a lightweight model that uses a common low-resolution camera to capture 2D skeletons from videos. Our model achieves excellent performance on multiple datasets covering a wide range of daily life activities and fall types.

\begin{figure*}[!t]
\centerline{\includegraphics[]{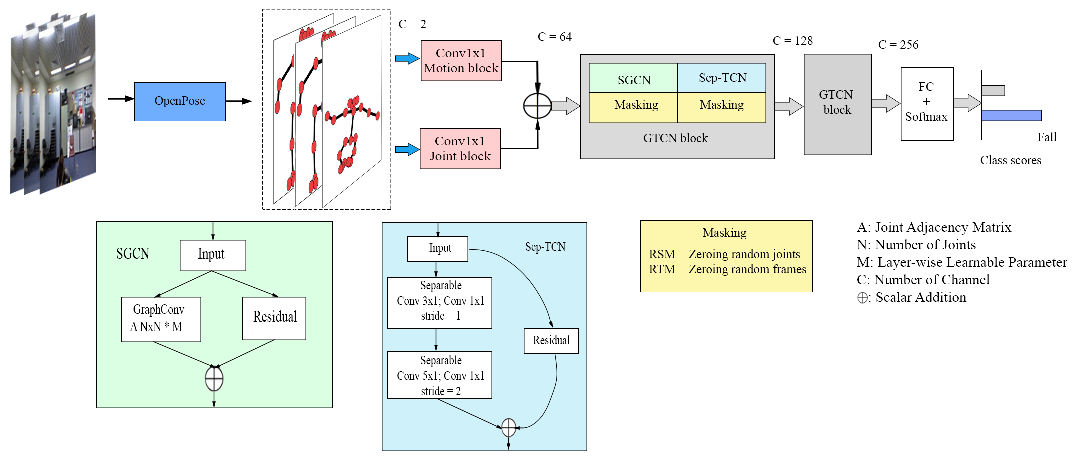}}
\caption{SDFA architecture: Firstly, skeleton joints are extracted from video streams using OpenPose (Section~\ref{openpose}). Then, linear projections of joint and motion streams are added to create a dynamic representation of the input skeletons (Section~\ref{input_stream}). The compact feature vector is then processed through spatial graph convolution (Section~\ref{sgcn_method}) and separable temporal convolution (Section~\ref{sep_tcn_method}) to encode local and global context aggregation over neighbouring joints and frames. Finally, global average pooling is performed over the encoded feature vector for classification.}
\label{method}
\end{figure*} 

%%%%%%%%%%%%%%%%%%%%%%%%%%%%%%%%%%%%%%%%%%
\section{Methodology}\label{Methodology}
Human skeleton joints can be connected by an undirected graph $G = (V,E)$ with $V$ joints constituting nodes and $E$ bones as edges. An adjacency matrix represents joint dependency, enabling powerful graph-based local and global feature aggregation. In this section, we present our proposed GCN based architecture that can: \textbf{(1)} create a dynamic representation of skeleton joints in a shared space, \textbf{(2)} capture the spatial relationship and \textbf{(3)} learn slow and fast temporal dynamics in an end-to-end learning framework. Figure~\ref{method} illustrates the proposed architecture.

\subsection{Skeleton Extraction using OpenPose}\label{openpose}
\begin{figure}[t]
\centerline{\includegraphics[width=\columnwidth]{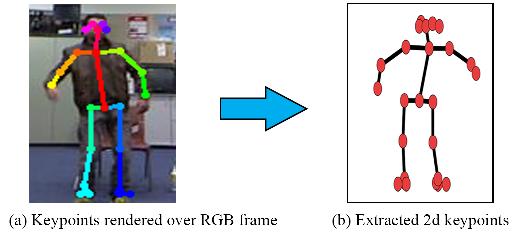}}
\caption{Skeleton extraction using OpenPose, (a) detected keypoints rendered over original video frame (b) extracted 2D joint coordinates.\label{fig:openpose}}
\end{figure}

OpenPose is a pose estimation algorithm that can detect keypoint positions from images or videos in real-time \cite{openpose2019}. It uses part affinity fields (PAF) to associate body parts to individuals to capture consistent skeletons over time for multiple persons. We used OpenPose to extract 2D joint coordinates from input video frames and the main part of the proposed architecture processes these coordinates. Figure~\ref{fig:openpose} represents the skeleton extraction process using OpenPose.

We restricted our videos to low resolution and a maximum of 30 fps frame rate for all datasets. Net resolution of OpenPose is set to $-1\times 256$ pixels to ensure execution on low power computational devices (-1 indicates arbitrary number of horizontal pixels). It directly extracts the joint coordinates and then discards the video since it is not required by our model during training or inference.  Thus, our proposed methodology ensures privacy as well as an inexpensive solution as it requires a low resolution camera.

\subsection{Early joint-motion stream fusion}\label{input_stream}
Skeleton joints provide valuable information about spatio-temporal relationships. Joint connections are consistent over time, and their change in spatial and temporal arrangements are discriminative indicators of actions. However, some activities such as \emph{Fall, Lying down, Squatting or Picking something up from the ground}, etc have analogous posture transitions. A crucial distinction among these activities is speed. Change in joint coordinate arrangements during \emph{Fall} happens within a very narrow window of time frame compared to other activities. Hence motion trajectory is added to the proposed architecture to enhance the distinction among action spaces.

Many human activity based models, such as MS-G3D \cite{liu2020disentangling}, exploit the representative power of motion using a two-stream architecture with parallel joint and motion streams. However, this doubles the overall complexity of the model. In contrast, we perform early fusion of motion and joint streams to achieve motion refined higher-level features without any significant increase in complexity. Unlike the conventional approach of adding final classification scores, both streams are processed through an encoding layer to project them onto a shared higher-dimensional space which provides a stable feature distribution by leveraging motion context in a more holistic way. The encoding layer contains batch normalization to stabilize input data followed by a $1\times1$ convolutional layer (Conv$1\times1$) which applies linear feature aggregation. Finally, both streams are combined by scalar addition resulting in a dynamic representation containing joint and motion traits.

\subsection{Spatial Graph Convolutional Network (SGCN)}\label{sgcn_method}
SGCN extracts spatial interaction between joints (green block in Figure~\ref{method}). Given an input skeleton, it aggregates relevant topological relationships from geometrically meaningful subgraphs. Convolution is guided using an adjacency matrix $A$ that represents skeleton graph connectivity between joints and bones. Separate convolutional modelling over a set of joints $B$ including both the current joint and its direct neighbours extracts corresponding relationships.We applied a learnable parameter $M$ at each layer to augment the adjacency matrix at higher layers with the increasing number of channels. This enhances joint association beyond physical connection and dynamically adjusts the matrix to input data. Spatial convolution is mathematically represented as
\begin{equation}\label{sgcn_eq} \tag{1}
f_{1}(v_{ti}) = \sum_{v_{tj}\in N(v_{ti})}\frac{1}{Z_{ti}(v_{tj})}\mathcal{F}\left(X(v_{tj})\right)W(l(v_{tj})),
\end{equation}

where $v_{tj}$ represents the $j^{th}$ dynamic motion infused joint at time $t$, $N(v_{ti})$ is the set of joints, $Z_{ti}(v_{tj})$ is the normalization term, $W(l(v_{tj}))$ is the weight matrix at layer $l$ for joint $v_{tj}$, $X(v_{tj})$ represents the skeleton frames and $\mathcal{F}(X(v_{tj}))$ represents the embeddings of the skeleton frames i.e. the summed output of Conv$1\times 1$ motion and joint blocks. Finally, neighboring joint matrix accumulation with the learned adjacency matrix is defined as

\begin{equation}\label{adjacency_eq} \tag{2}
f_{out}(v_{ti}) = \sum_{v_{tj}\in B(v_{ti})}A(v_{tj})M_l(v_{tj})f_{1}(v_{tj})
\end{equation}

where $B(v_{ti})$ is the set of neighbouring joints, $A(v_{tj})$ is the adjacency matrix, $M_l(v_{tj})$ is the learned matrix at layer $l$ and $f_{1}(v_{tj})$ is the $v_{tj}$ joint of the output matrix from Eq~\ref{sgcn_eq}. The weight sharing technique of SGCN among neighbouring joints enables it to capture intricate spatial dependencies. We add a residual connection after SGCN. Unlike standard residual connection where either an identity function or a $1\times 1$ Conv is used to retain the original feature vector, we applied spatial max pooling which reinforces the most active joints and thus adaptively increases spatial focus. 

\subsection{Sep-TCN: Separable Temporal Convolutional Network}\label{sep_tcn_method}
Traditional convolution operation performs $K\times K\times C$ multiplications every time the kernel moves which significantly increases the total FLOPS. Depthwise separable convolution decomposes this process into depthwise (DW) and pointwise (PW) convolution, where DW applies a $K\times K\times 1$ filter and PW applies a $1\times 1\times C$ filter \cite{MobileNets2017}. This dramatically reduces the total number of transpose operations needed during convolution. Hence, the number of model parameters is greatly reduced without affecting performance and the model runs more than twice as fast as its traditional counterpart \cite{MobileNets2017}. 

\begin{figure}[!t]
\centerline{\includegraphics[width=\columnwidth]{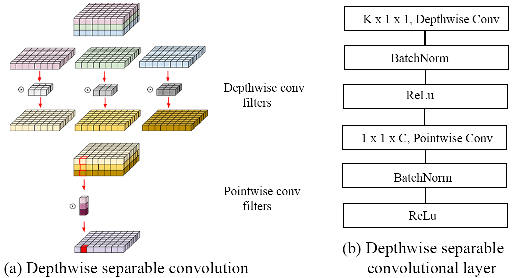}}
\caption{Depthwise separable convolution: (a) filtering operations split into two, depthwise and pointwise convolution. (b) represents a basic depthwise separable convolution layer used in the proposed model \cite{MobileNets2017}.\label{septem}}
\end{figure}

We applied a kernel of $3\times1$ with stride 1 for the first Sep-TCN and $5\times1$ with stride 2 for the second Sep-TCN layer, resulting in a filter size of $3 \times 1 \times 1$ and $5 \times 1 \times 1$ for DW. It is indicated as $K \times 1 \times 1$ in Figure 3(b). We kept spatial kernel dimension 1 instead of $K$ as in [23] to retain the original joint dimension throughout the computation. Thus, our temporal module first searches in local temporal proximity followed by a wider space. Figure~\ref{septem} represents the filtering approach of DW and PW, and a basic Sep-TCN layer. 
To enhance temporal focus, we applied a residual connection using temporal max pooling. It concentrates the dominant frames in the feature vector. 

\subsection{Randomized Masking}
To tackle the problem of model overfitting, we implemented randomized masking over joints and frames which forces our model to learn from sparse feature matrices. Our implementation is inspired from DropGraph \cite{cheng2020eccv} which implements the functionality of dropout layers. Similar to dropping layers, it drops a block of consecutive frames and several joints. However, variation in skeleton postures in consecutive frames are negligible and dropping them does not affect the overall activity pattern significantly. Alternatively, disjoint frames at different time points carry important details and dropping them enables the model to learn more generalized features. 

\begin{figure}[!t]
\centerline{\includegraphics[width=\columnwidth]{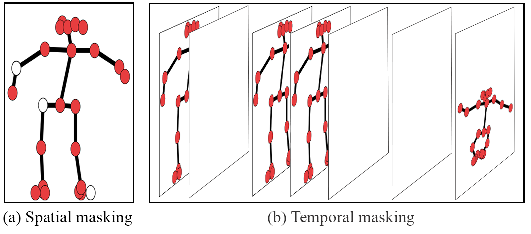}}
\caption{Randomized masking. (a) Spatial joints (the three white ones) are masked. (b) Temporal frames (number 2, 5, and 6) are masked.\label{masking}}
\end{figure}

Therefore, we sparsify our input skeleton sequences by randomly masking frames and joints at different positions with a heuristically chosen probability, used during training of the model. Experiments indicate that this variant better generalizes as it ensures learning of important cues in a constrained manner and achieves better results. During testing, we set masking probability to 0 and use all available data. Figure~\ref{masking} represents a sample example of randomized spatial and temporal masking.

%%%%%%%%%%%%%%%%%%%%%%%%%%%%%%%%%%%%%%%%%%
\section{Experiments}\label{Experiments}
We present experimental results on five fall detection datasets using various evaluation settings and compare our method with existing state-of-the-art to demonstrate the excellence of our architectural design.

\subsection{Datasets}\label{datasets}
Most of the literature use fall detection datasets that have only 4-6 types of non-fall actions, which do not cover actions from a realistic daily life scenario. Therefore, we included UWA3D \cite{Rahmani_tpami_2016}, NTU-60 \cite{ShahroudyAmir2016NRAL} and NTU-120 \cite{Liu_2019_NTURGBD120} datasets that provide a more comprehensive way of evaluation with a much wider range of daily life activities including \emph{Fall}. Besides, they are collected from a wide range of subjects and viewpoint variations compared to previous datasets. Table~\ref{tab_datasets} represents an overview of the datasets used in this work.

\begin{table}[!hbt]
\caption{Fall detection dataset details. Activities of Daily Living (ADL) are non-fall actions}. \label{tab_datasets}
\setlength{\tabcolsep}{3pt}
\setlength\extrarowheight{1pt}
\begin{tabular}{|p{50pt}|P{30pt}|P{25pt}|P{35pt}|P{38pt}|P{22pt}|}
\hline
\textbf{Dataset}	& \textbf{Subjects}	& \textbf{Views}	& \textbf{Fall types}	& \textbf{ADL types} & \textbf{Year}\\
\hline
URFD \cite{UR_fall_2014} & 5 & 2 & 3  & 6 & 2014\\
UPFD \cite{UP_Fall_2019} & 17 & 2 & 5 & 6 & 2019\\
UWA3D \cite{Rahmani_tpami_2016} & 10 & 4 &	1 & 29 & 2015\\
NTU-60 \cite{ShahroudyAmir2016NRAL}	& 40 & 80 &	1 & 59 & 2016\\
NTU-120 \cite{Liu_2019_NTURGBD120} & 106 & 155 & 1 & 119 & 2019\\
\hline
\end{tabular}
\end{table}

\textbf{URFD:} UR fall detection dataset has 60 \emph{Fall} samples captured from 2 camera viewpoints and 40 non-fall samples from one camera viewpoint. Videos are captured  at $640\times 480$ resolution and at 30 fps \cite{UR_fall_2014}.

\textbf{UPFD:} UP-fall detection dataset has 1118 samples. It provides a more challenging scenario with five different fall types: \emph{Falling forward using hands}, \emph{Falling forward using knees}, \emph{Falling backwards}, \emph{Falling sitting in empty chair} and \emph{Falling sideways}. Particularly, \emph{Falling sitting in empty chair} is quite complex as the person assumes the posture of sitting on a chair before hitting the ground \cite{UP_Fall_2019}. Videos are captured  at $640\times 480$ resolution and at 18 fps.

\begin{figure}[!htbp]
\centerline{\includegraphics[width=\columnwidth]{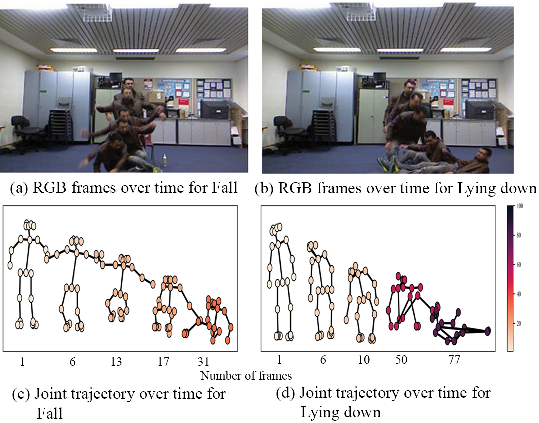}}
\caption{Samples of \emph{Fall} and \emph{Lying down} activities from UWA3D dataset. (a)-(b) are superimposed RGB frames and (c)-(d) are corresponding skeleton joint frames. As the colour indicates, both activities have similar postures differing only in their temporal occurrence. Similar skeleton poses for \emph{Lying down} are spread out further over time indicating longer duration and moderate transition speed (deeper colour indicates higher frame number, therefore a later time of occurrence).} \label{fig:rgb_skel}
\end{figure}

\textbf{UWA3D:} UWA3D Multiview Activity II dataset has 30 actions including \emph{Lying down} which significantly overlaps with \emph{Fall} (see Figure \ref{fig:rgb_skel}). Actions are performed without a break and hence the same action has different start and end body pose which makes the data more realistic and challenging \cite{Rahmani_tpami_2016}. Videos are captured  at $320\times 240$ resolution and at 30 fps.

\textbf{NTU-60} has 56,880 video samples and is divided into cross-view and cross-subject action recognition tasks \cite{ShahroudyAmir2016NRAL}. Only one out of the 60 actions is fall action. Videos are captured at $1920\times 1080$ resolution and at 30 fps. We reduced the resolution to $256\times 256$ for efficiency.

\textbf{NTU-120} is the extended version of NTU-60. It has 114,480 samples and 60 additional activities including \emph{Squatting down} which is very similar to \emph{Fall}. It provides the largest number of participants and viewpoints \cite{Liu_2019_NTURGBD120}. It has similar fps and resolution to NTU-60 which was reduced to $256\times 256$ in our experiments.

\subsection{Data Preprocessing}\label{data_preprocessing}
Since some videos contain no humans in the beginning, OpenPose produces empty frames with zero values. Our preprocessing module detects and discards these frames automatically. Zero values in the input frames do not activate their corresponding convolutional unit in the next layers and thus do not affect classification accuracy. However, not removing them increases training time. By removing such frames, we achieve faster convergence. Besides, OpenPose sometimes mistakenly detects furniture, such as a chair, as a human skeleton. These inanimate skeleton-like structures have a much lower overall standard deviation than humans. Therefore, we only kept valid skeleton frames with the highest standard deviation in the temporal dimension.
After data cleaning, all sequences were restricted to 300 frames, repeating frames where necessary. For the UPFD dataset, the sequence length was set to 1145 since those sequences are much longer. Next, view-invariant transformation is performed to achieve similar skeleton orientations by transforming the spine joints to the origin. Finally, we performed data normalization to maintain zero mean and unit variance.

\subsection{Implementation Details}
We used the same hyperparameter settings for evaluating on all datasets. Specifically, an SGD optimizer with 0.9 momentum and 0.01 initial learning rate was used to train the model. Each model was trained for 50 epochs with learning rate decreasing by 10\% every 10th epoch. We used six different evaluation types to analyze the efficiency of our model. Table \ref{model_specs} demonstrates a quantitative comparison of our model with other state-of-the-art methods in terms of computational complexity (Floating Point Operations Per Second (FLOPS)), number of parameters and inference time. We could not compare some methods \cite{Carlier2020FallDA, ESPINOSA2019103520, Tsai_2019} as their code is not publicly available nor are their architectural specifications to calculate the computational complexity of their models.

Conventionally, a random training-test split is used to evaluate classification algorithms. However, human fall detection in videos has some unique characteristics. A person may appear in very diverse viewing angles w.r.t. the camera. This results in very different appearances of the human as well as self occlusions (partial side views). The cross-view evaluation setup uses samples captured from viewpoints of 2 cameras for training and samples captured from different viewpoints of a separate camera for testing. Such an experimental setup validates the method's robustness to viewpoint variations and self occlusions. Another unique characteristic of the human fall detection problem is that due to physical disparities, people have distinct motion characteristics. Hence, it is crucial to validate a fall detection system for previously unseen people. Cross-subject evaluation setting ensures that the persons that appear in the training set are disjoint from those who appear in the test set.

\begin{table}
\caption{Comparison of our model with different fall detection methods.\label{model_specs}}
\setlength{\tabcolsep}{3pt}
\setlength\extrarowheight{1pt}
\begin{tabular}{|p{85pt}|P{30pt}|P{55pt}|P{52pt}|}
\hline
\textbf{Method}	& \textbf{FLOPS (G)}	& \textbf{Number of parameters (M)}	& \textbf{Inference time (ms)}\\
\hline
Inception-ResNet-V2 \cite{XU2020123205} & 39.58 & 56.0 & 25.9 \\
AlexNet \cite{Tsai_2019} & 0.725 & 61.0 & 1.28 \\
ST-GCN \cite{Keskes_2021} & 24.59 & 4.38 & 9.95 \\
GCN \cite{sania2021} & 16.17 & 1.26 & 1.32 \\
\hline
\textbf{STDF Net} & \textbf{1.15} & \textbf{0.34} & \textbf{1.05 }\\
\hline
\end{tabular}
\end{table}

\subsection{Results}
Table \ref{results} presents experimental results on all five datasets: UR Fall, UP Fall, UWA3D, NTU-60 and NTU-120 with the following evaluation settings, 5-fold cross-validation with Cross-Fall and Cross-Trial (UP Fall), Cross-View (UWA3D and NTU-60), Cross-Subject (NTU-60 and NTU 120) and Cross-Setup (NTU 120). We prepared training and testing data considering the subject, viewpoint, setup and fall type variations, which enable us to test our model against these variations. For the URFD dataset, we followed the standard 70-30\% train-test split as suggested in a study on URFD \cite{Chen2020}. UPFD dataset has five different types of \emph{Fall}. Therefore, we followed standard cross-trial and used a new cross-fall evaluation. In cross-trial, we used trial 1 and 2 data for training, and trial 3 data for testing. In cross-fall, four types of falls are used to train and the remaining unseen fall type is used for testing. We followed 5 fold cross-validation in a leave-one-out fashion where one fall type is set aside for testing. Individual and average results are reported. 

For NTU-60/120 datasets, we used cross-subject, cross-view and cross-setup evaluation settings. NTU-60 and NTU-120 cross-subject use 20 and 53 subjects respectively for training and the remaining for testing respectively. Both NTU-60/120 uses camera view 2 and 3 for training and 1 for testing in cross-view evaluation setting. Cross-setup uses even-numbered samples for training and odd for testing.

\begin{table*} [!hbt]
\caption{Results on five datasets. All values are in percentage}. \label{results}
\setlength{\tabcolsep}{3pt}
\setlength\extrarowheight{1pt}
\centering
\begin{tabular}{|P{35pt}|P{50pt}|P{37pt}|P{30pt}|P{35pt}|P{30pt}|P{45pt}|P{22pt}|P{32pt}|}
\hline
\textbf{Dataset}& \textbf{Evaluation}& \textbf{Specificity}& \textbf{Recall} & \textbf{Precision} & \textbf{FP Rate} & \textbf{F1 Measure} &  \textbf{AUC} & \textbf{Accuracy}\\
\hline
URFD	& 70-30 & 100 & 100 & 100 & 0 & 100 & 100 & 100\\ 
\hline

\multirow{6}{*}{UPFD} & Cross-fall$_1$ & 86.81 & 88.24 & 78.95 & 13.19 & 83.33 & 87.52 & 87.32\\

 & Cross-fall$_2$ & 82.97 & 96.08 & 75.97 & 17.03 & 84.85 & 89.52 & 87.68\\

 & Cross-fall$_3$ & 84.07 & 94.12 & 76.80 & 15.93 & 84.58 & 89.09 & 87.68 \\

 & Cross-fall$_4$ & 87.36 & 98.04 & 81.30 & 12.64 & 88.89 & 92.70 & 91.19 \\

 & Cross-fall$_5$ & 80.77 & 90.19 & 72.44 & 19.23 & 80.35 & 85.48 & 84.16 \\

 & 5-fold (avg) & 84.40 & 93.33 & 77.09 & 15.60 & 84.40 & 88.86 & 87.61 \\

 & Cross-trial & 85.15 & 92.94 & 84.04 & 14.85 & 88.27 & 89.04 & 88.71 \\
\hline

UWA3D & Cross-view & 99.23 & 88.89 & 80.00 & 0.77 & 84.21 & 94.06 & 98.88 \\
\hline

\multirow{2}{*}{NTU-60} & Cross-subject & 99.80 & 82.61 & 87.69 & 0.19 & 85.08 & 91.21 & 99.51\\

 & Cross-view & 99.70 & 91.14 & 83.97 & 0.29 & 87.41 & 95.42 & 99.56\\ 
\hline

\multirow{2}{*}{NTU-120} & Cross-subject & 99.35 & 85.51 & 41.99 & 0.65 & 56.33 & 92.43 & 99.27\\ 
 & Cross-setup & 99.66 & 73.22 & 65.07 & 0.34 & 68.90 & 86.44 & 99.44 \\

\hline
\end{tabular}
%\vspace{-1em}
\end{table*}

%%%%%%%%%%%%%%%%%%%%%%%%%%%%%%%%%%%%%%%%%%
Table \ref{sota_results} compares our proposed model with state-of-the-art results. Our model outperforms most works except \cite{sania2021} on NTU-60/120. However, \cite{sania2021} requires 3D skeletons extracted with the Kinect (or similar active) sensor. Compared to conventional cameras, Kinect is more expensive, bulky and consumes more energy being an active sensor. Our method requires a simple camera, facilitating an effective and cheaper solution. Moreover, our model has 14 times (1.15 vs 16.17 GFLOPS) lower computational complexity and 0.92 Million fewer parameters compared to \cite{sania2021}, ensuring the fastest inference time (about 20\% less than \cite{sania2021}). Note that [15][18] only used a {\em selected} subset of the NTU datasets and a simpler evaluation criteria. Our method still outperforms [15][18] even though we evaluate on the full NTU datasets and using much more challenging scenarios of cross-subject and cross-view.

\begin{table*}[!hbt]
\caption{Comparison of proposed fall detection model against existing works. All values are in percentage. \label{sota_results}}
\setlength{\tabcolsep}{3pt}
\setlength\extrarowheight{1.2pt}
\centering
\begin{tabular}{|p{90pt}|P{60pt}|P{85pt}|P{40pt}|P{40pt}|P{40pt}|P{40pt}|P{40pt}|P{22pt}|}
\hline
\textbf{Model}& \textbf{Modality} & \textbf{Evaluation}& \textbf{URFall} & \textbf{UPFall} & \textbf{UWA3D} & \textbf{NTU-60} & \textbf{NTU-120} & \textbf{Year}\\
\hline

1D-CNN \cite{Tsai_2019} & RGB $\rightarrow$ 2D Skel & \makecell{70-30 (subset)} & Acc 98 & - & - & Acc 99.20 & - & 2019 \\
\hline

AlexNet \cite{Tsai_2019} & RGB $\rightarrow$ 2D Skel & \makecell{70-30 (subset)} & - & - & - & Acc 98.90 & - & 2019 \\
\hline

Handcrafted features with MLP \cite{Wang2020} & RGB $\rightarrow$ 2D Skel & \makecell{-} & \makecell{F1 97.78 \\Acc 97.33} & - & - & - & - & 2020 \\
\hline

CNN with dense flow \cite{Carlier2020FallDA} & RGB & \makecell{5 fold cross-validation} & \makecell{Se 86.2\\F1 87.29
\\FP 11.6} & - & - & - & - & 2020 \\
\hline

CNN \cite{ESPINOSA2019103520} & RGB & \makecell{-} & \makecell{-} & \makecell{F1 72.94\\Acc 82.26} & - & - & - & 2020 \\
\hline

Inception-ResNet-V2 \cite{XU2020123205} & RGB $\rightarrow$ 2D Skel & - & Acc 91.70  & - & - &  Acc 91.70  & - & 2020 \\
\hline

ST-GCN \cite{Keskes_2021} & 3D Skel & Cross-subject (subset) & - & - & - & Acc 92.91 & - & 2021 \\
\hline

GCN \cite{sania2021} & 3D Skel & \makecell{70-30, Cross-subject,\\Cross-setup, Cross-view} & - & - & \makecell{AUC 83.13\\Acc 98.40\\\textbf{FP 0.42}} & \makecell{\textbf{AUC 99.03}\\\textbf{Acc 99.93}\\\textbf{FP 0.04}} & \makecell{\textbf{AUC 96.12}\\\textbf{Acc 99.83}\\\textbf{FP 0.13}} & 2021\\
\hline

\textbf{SDFA Net} & RGB $\rightarrow$ 2D Skel & \makecell{\textbf{70-30, Cross-subject}\\\textbf{Cross-setup, Cross-view}\\\textbf{Cross-trial, Cross-fall$_x$}} & \makecell{\textbf{AUC 100}\\\textbf{Acc 100}\\ \textbf{FP 0.0}} & \makecell{\textbf{AUC 89.04}\\\textbf{Acc 88.71}\\\textbf{FP 14.85}} & \makecell{\textbf{AUC 94.06}\\ \textbf{Acc 98.88}\\ FP 0.77} & \makecell{AUC 95.42\\Acc 99.56\\FP 0.29} & \makecell{AUC 92.43\\Acc 99.27\\FP 0.65} & \\
\hline
\end{tabular}
\end{table*}

%%%%%%%%%%%%%%%%%%%%%%%%%%%%%%%%%%%%%%%%%%
\subsection{Ablation Studies}\label{Ablation_Studies}
We analyzed individual components of our model to investigate their contribution to the overall performance. We started with the baseline model followed by different adjustments to improve performance and architectural explainability. All experiments have been done on the UWA3D dataset in cross-view setting. Table \ref{ablation} represents results from our ablation studies with different network modules in terms of accuracy and area under the curve (AUC).

\begin{table}[!hbt]
\caption{Accuracy and AUC measure with individual modules on UWA3D dataset. All values are in percentage.\label{ablation}}
\setlength{\tabcolsep}{3pt}
\setlength\extrarowheight{1pt}
\centering
\begin{tabular}{|p{140pt}|P{35pt}|P{22pt}|}
\hline
\textbf{Model configuration} & \textbf{Accuracy} & \textbf{AUC}\\
\hline
Baseline & 96.28 & 55.17\\

Learnable adjacency matrix & 98.14 & 88.31\\
Block temporal dropgraph & 97.40 & 66.47\\
Randomized spatio-temporal masking & 98.88 & 94.06\\
\hline
\end{tabular}
\end{table}

\subsubsection{Baseline Model} Our baseline model includes two layers of SGCN followed by one traditional temporal convolutional network. We used a $25\times 25$ adjacency matrix for neighbourhood contextual aggregation and a single TCN with stride = 2. It achieved 96.28\% accuracy with 55.17\% AUC. However, the lower AUC indicates that the model is not powerful enough to capture the complex spatio-temporal dependency in skeleton graphs to distinguish falls. Therefore, we investigated the following procedures to improve the performance and generalization of our proposed method.

\subsubsection{Learnable Adjacency Matrix} A conventional $25\times 25$ adjacency matrix only depicts one hop neighbourhood relationships which does not impart any knowledge about distant joint dependency in contrast to a three hop adjacency matrix of shape $3\times 25\times 25$ as suggested in \cite{sania2021}. Although a reduced shape dramatically decreases the number of model parameters and FLOPS, it also sacrifices information. Moreover, spatio-temporal representation is not straightforward in the higher layer, and a static $25\times 25$ matrix restricts the learning capability. To alleviate this constraint, we introduce a layer-wise learnable parameter $\mu$ that ensures a data and layer adaptive dynamic adjacency matrix. With the enhanced learning capability, our model achieved 98.14\% accuracy with  88.31\% AUC.

\subsubsection{Randomized Masking} To further boost the performance, we added various regularization techniques such as dropout layers, L2 regularization etc. These techniques penalize learning and help to capture more randomness, thus reducing the chances of overfitting. However, for graph convolution, dropgraph \cite{cheng2020eccv} is a more intuitive choice as it penalizes joint focus. To explore its efficacy in our model, we first used the block temporal dropgraph method from \cite{cheng2020eccv} after each layer. Despite its apparent potential, this implementation decreased our accuracy from 98.14\% to 97.40\% which indicates that dropping a block of frames is not effective since consecutive frames contain inadequate posture variations that do not contribute any significant information. Thus, the model fails to achieve sufficient generalization power. 

Therefore, we propose randomized spatial and temporal masking, forcing the model not to favour any specific set of joints or frames. With this modification, our model achieved 98.88\% accuracy with 94.06\% AUC. We investigated the use of different variations of masking layers. Experiments indicate that a single masking layer after each convolutional layer, excluding the initial input stream fusion layer, gives the best performance without any significant increase in the number of model parameters.

\subsubsection{Early Fusion of Joint and Motion Stream} The intuition behind using joint and motion streams is to magnify motion variation. As shown in Figure~\ref{fig:rgb_skel}, \emph{Fall} and \emph{Lying down} have similar trajectories except the speed of progression. Therefore, merging joint and motion trajectories should better differentiate these events. On the other hand, a parallel two-stream architecture will almost double the computational complexity and memory requirements. Hence, we propose early fusion that provides similar benefits without increasing the model size. Table \ref{embed_results} shows the results using different streams. As anticipated, combining the streams increases recall and F1 measure, two of the most crucial evaluation metrics. Increased recall ensures that events that are actually \emph{Fall}, are detected at a higher rate.

\begin{table*}[!htb]
\caption{Experimental results on UWA3D dataset with joint, motion and combined joint-motion streams \label{embed_results}}
\setlength{\tabcolsep}{3pt}
\setlength\extrarowheight{1pt}
\footnotesize
\centering
\begin{tabular}{|P{60pt}|P{60pt}|P{35pt}|P{45pt}|P{45pt}|P{60pt}|P{30pt}|P{50pt}|}
\hline
\textbf{Stream}& \textbf{Specificity}& \textbf{Recall} & \textbf{Precision} & \textbf{FP Rate} & \textbf{F1 Measure} &  \textbf{AUC} & \textbf{Accuracy}\\
\hline
Joint & 100 & 55.56 & 100 & 0 & 71.43 & 77.79 & 98.51 \\
\hline
Motion & 100 & 55.56 & 100 & 0 & 71.43 & 77.78 & 98.51 \\
\hline
Joint-motion & 99.23 & 88.89 & 80.00 & 0.77 & 84.21 & 94.06 & 98.88 \\
\hline
\end{tabular}
\end{table*}

%%%%%%%%%%%%%%%%%%%%%%%%%%%%%%%%%%%%%%%%%%
\section{Discussion}\label{Discussion}
Our proposed model has an order of magnitude lower computational complexity (in terms of FLOPS). Our model also has fewer (1/4th) parameters compared to existing architectures. This reduces hardware requirements in terms of processing power and memory, leading to lower cost. Reduced computational complexity also results in faster run time as shown in Table \ref{model_specs}. Our method takes 0.27 ms less time (about 20\% faster) than the best competing method. Moreover, our model uses 2D skeletons extracted from images captured with conventional cameras and does not require active sensors like Kinect, which consume more power and are relatively more expensive.

A fall detection technique must be able to generalize to different situations. The best way to validate this is to see if it achieves comparative performance across multiple different datasets with changing viewpoints, surrounding environments, subjects and types of activities. Though most studies report great accuracy, they provide evaluations on one or two datasets that includes only a few types of activities. To overcome concerns about generalizability of the solution, we incorporated five different datasets in our study with a wide variety of subjects and activities. It enabled us to conduct a thorough investigation of our model. 

Moreover, our evaluation settings provide a comprehensive approach to analyze factors in model learning. Instead of the conventional 70\%-30\% splits used in existing works, we followed a rigorous set of evaluations to consider different data characteristics. Each evaluation setting examines specific key factors. For example, cross-subject analyzes model efficiency by training with samples from one set of subjects and testing from a different set of subjects. Similarly, cross-view investigates the model's viewpoint invariance capability. Table \ref{results} summarises the results on all five datasets with six types of evaluation settings. We can see that cross-view and cross setup evaluation perform slightly better than cross subject as subject related variations are more complex. 

We also observe that false-positive rates for the UPFD dataset are much higher compared to other datasets. UPFD provides a greater challenge with two very similar activities: \emph{Falling sitting in empty chair} and \emph{Sitting}. Besides, for the activity \emph{Laying}, there is no motion at all as the subject remains stationary on a mattress. These activities contribute to the comparatively higher false positives. In addition, UPFD has only 1118 samples with many frames being empty. This further complicates proper training. Though NTU datasets also contain similar activities, the huge number of samples enables the model to learn to distinguish these activities.

%%%%%%%%%%%%%%%%%%%%%%%%%%%%%%%%%%%%%%%%%%
\section{Conclusion}\label{Conclusions}
In this paper, we presented a lightweight fall detection system with graph-based convolution. Our proposed system is capable of effectively detecting falls from 2D skeleton data, hence, eliminating the need to store the videos or use expensive sensors. We derive a powerful feature aggregation technique with discriminative local and global spatio-temporal focus. Besides, early fusion of motion and joint stream enhances feature representation without over-burdening the network. Thus, our proposed architecture enjoys low computational complexity as well as the advantage of precise attention of graph convolution. With extensive experiments on five large scale datasets, our model outperforms existing models in various performance matrices. 

%%%%%%%%%%%%%%%%%%%%%%%%%%%%%%%%%%%%%%%%%%

\bibliographystyle{IEEEtran}
\bibliography{bibtex}

\begin{thebibliography}{10}
\providecommand{\url}[1]{#1}
\csname url@samestyle\endcsname
\providecommand{\newblock}{\relax}
\providecommand{\bibinfo}[2]{#2}
\providecommand{\BIBentrySTDinterwordspacing}{\spaceskip=0pt\relax}
\providecommand{\BIBentryALTinterwordstretchfactor}{4}
\providecommand{\BIBentryALTinterwordspacing}{\spaceskip=\fontdimen2\font plus
\BIBentryALTinterwordstretchfactor\fontdimen3\font minus
  \fontdimen4\font\relax}
\providecommand{\BIBforeignlanguage}[2]{{%
\expandafter\ifx\csname l@#1\endcsname\relax
\typeout{** WARNING: IEEEtran.bst: No hyphenation pattern has been}%
\typeout{** loaded for the language `#1'. Using the pattern for}%
\typeout{** the default language instead.}%
\else
\language=\csname l@#1\endcsname
\fi
#2}}
\providecommand{\BIBdecl}{\relax}
\BIBdecl

\bibitem{MATHON2017e50}
C.~Mathon, F.~Beaucamp, F.~Roca, P.~Chassagne, A.~Thevenon, and F.~Puisieux,
  ``Post-fall syndrome: Profile and outcomes,'' \emph{Annals of Physical and
  Rehabilitation Medicine}, 2017, 32nd Annual Congress of the French Society of
  Physical and Rehabilitation Medicine.

\bibitem{disease_2021}
M.~Immonen, M.~Haapea, H.~Similä, H.~Enwald, N.~Keränen, M.~Kangas,
  T.~Jämsä, and R.~Korpelainen, ``Association between chronic diseases and
  falls among a sample of older people in finland,'' \emph{BMC Geriatrics},
  2020.

\bibitem{WHO_FALL}
W.~H. Organization, ``Falls,'' 26-Apr-2021 [Online],
  \url{www.who.int/news-room/fact-sheets/detail/falls}, [Accessed:
  28-June-2021].

\bibitem{UN_age}
U.~Nations, ``Department of {E}conomic and {S}ocial {A}ffairs, {U}nited
  {N}ations,'' 2017 [Online],
  \url{www.un.org/en/development/desa/population/publications/pdf/ageing/\\WPA2017\_Highlights.pdf},
  [Accessed: 28-June-2021].

\bibitem{Alarifi_2021}
A.~Alarifi and A.~Alwadain, ``Killer heuristic optimized convolution neural
  network-based fall detection with wearable {IoT} sensor devices,''
  \emph{Measurement}, 2021.

\bibitem{Carlier2020FallDA}
A.~Carlier, P.~Peyramaure, K.~Favre, and M.~Pressigout, ``Fall detector adapted
  to nursing home needs through an optical-flow based {CNN},'' \emph{2020 42nd
  Annual International Conference of the IEEE Engineering in Medicine \&
  Biology Society (EMBC)}, 2020.

\bibitem{Liu_ICAIBD}
J.-x. Liu, R.~Tan, N.~Sun, G.~Han, and X.-f. Li, ``Fall detection under privacy
  protection using multi-layer compressed sensing,'' in \emph{2020 3rd
  International Conference on Artificial Intelligence and Big Data (ICAIBD)},
  2020.

\bibitem{Medical_alert_2021}
M.~E. Team, ``{F}all {D}etection {F}alse {A}larms in {S}mart {W}atches,''
  3-Jul.-2021 [Online],
  \url{https://www.medicalalertbuyersguide.org/articles/fall-detection-false-alarms-in-smart-watches/},
  [Accessed: 23-Nov.-2021].

\bibitem{dimentia_fall}
E.~Heerema, ``Common {C}auses of {F}alls in {P}eople {W}ith {D}ementia,''
  20-Apr.-2022 [Online],
  \url{https://www.verywellhealth.com/causes-of-falls-in-people-with-dementia-98558},
  [Accessed: 20-Apr.-2022].

\bibitem{ASLAN20151023}
M.~Aslan, A.~Sengur, Y.~Xiao, H.~Wang, M.~C. Ince, and X.~Ma, ``Shape feature
  encoding via fisher vector for efficient fall detection in depth-videos,''
  \emph{Applied Soft Computing}, 2015.

\bibitem{bian2015}
Z.-P. Bian, J.~Hou, L.-P. Chau, and N.~Magnenat-Thalmann, ``Fall detection
  based on body part tracking using a depth camera,'' \emph{IEEE Journal of
  Biomedical and Health Informatics}, 2015.

\bibitem{Chen2020}
Y.~Chen, W.~Li, L.~Wang, J.~Hu, and M.~Ye, ``Vision-based fall event detection
  in complex background using attention guided bi-directional lstm,''
  \emph{IEEE Access}, 2020.

\bibitem{Kottari2020}
K.~N. Kottari, K.~K. Delibasis, and I.~G. Maglogiannis, ``Real-time fall
  detection using uncalibrated fisheye cameras,'' \emph{IEEE Transactions on
  Cognitive and Developmental Systems}, 2020.

\bibitem{liu2020disentangling}
Z.~Liu, H.~Zhang, Z.~Chen, Z.~Wang, and W.~Ouyang, ``Disentangling and unifying
  graph convolutions for skeleton-based action recognition,'' in
  \emph{Proceedings of the IEEE/CVF Conference on Computer Vision and Pattern
  Recognition}, 2020.

\bibitem{Keskes_2021}
O.~Keskes and R.~Noumeir, ``Vision-based fall detection using {ST-GCN},''
  \emph{IEEE Access}, 2021.

\bibitem{Wang2020}
B.-H. Wang, J.~Yu, K.~Wang, X.-Y. Bao, and K.-M. Mao, ``Fall detection based on
  dual-channel feature integration,'' \emph{IEEE Access}, 2020.

\bibitem{XU2020123205}
Q.~Xu, G.~Huang, M.~Yu, and Y.~Guo, ``Fall prediction based on key points of
  human bones,'' \emph{Physica A: Statistical Mechanics and its Applications},
  2020.

\bibitem{Tsai_2019}
T.-H. Tsai and C.-W. Hsu, ``Implementation of fall detection system based on
  3{D} skeleton for deep learning technique,'' \emph{IEEE Access}, 2019.

\bibitem{openpose2019}
Z.~{Cao}, G.~{Hidalgo Martinez}, T.~{Simon}, S.~{Wei}, and Y.~A. {Sheikh},
  ``Openpose: Realtime multi-person 2d pose estimation using part affinity
  fields,'' \emph{IEEE Transactions on Pattern Analysis and Machine
  Intelligence}, 2019.

\bibitem{Nahian_access}
M.~J.~A. Nahian, T.~Ghosh, M.~H.~A. Banna, M.~A. Aseeri, M.~N. Uddin, M.~R.
  Ahmed, M.~Mahmud, and M.~S. Kaiser, ``Towards an accelerometer-based elderly
  fall detection system using cross-disciplinary time series features,''
  \emph{IEEE Access}, 2021.

\bibitem{MA201944}
C.~Ma, A.~Shimada, H.~Uchiyama, H.~Nagahara, and R.~ichiro Taniguchi, ``Fall
  detection using optical level anonymous image sensing system,'' \emph{Optics
  \& Laser Technology}, 2019, special Issue: Optical Imaging for Extreme
  Environment.

\bibitem{sania2021}
S.~Zahan, G.~M. Hassan, and A.~Mian, ``Modeling human skeleton joint dynamics
  for fall detection,'' in \emph{2021 Digital Image Computing: Techniques and
  Applications (DICTA)}, 2021.

\bibitem{MobileNets2017}
A.~G. Howard, M.~Zhu, B.~Chen, D.~Kalenichenko, W.~Wang, T.~Weyand,
  M.~Andreetto, and H.~Adam, ``Mobilenets: Efficient convolutional neural
  networks for mobile vision applications,'' \emph{CoRR}, 2017.

\bibitem{cheng2020eccv}
K.~Cheng, Y.~Zhang, C.~Cao, L.~Shi, J.~Cheng, and H.~Lu, ``Decoupling gcn with
  dropgraph module for skeleton-based action recognition,'' in
  \emph{Proceedings of the European Conference on Computer Vision (ECCV)},
  2020.

\bibitem{Rahmani_tpami_2016}
H.~Rahmani, A.~Mahmood, D.~Huynh, and A.~Mian, ``Histogram of oriented
  principal components for cross-view action recognition,'' \emph{IEEE
  Transactions on Pattern Analysis and Machine Intelligence}, 2016.

\bibitem{ShahroudyAmir2016NRAL}
A.~Shahroudy, J.~Liu, T.-T. Ng, and G.~Wang, ``{NTU RGB+D}: A large scale
  dataset for 3d human activity analysis,'' in \emph{The IEEE Conference on
  CVPR}, June 2016.

\bibitem{Liu_2019_NTURGBD120}
J.~Liu, A.~Shahroudy, M.~Perez, G.~Wang, L.-Y. Duan, and A.~C. Kot, ``{NTU
  RGB+D 120}: A large-scale benchmark for 3d human activity understanding,''
  \emph{IEEE Transactions on Pattern Analysis and Machine Intelligence}, 2019.

\bibitem{UR_fall_2014}
B.~Kwolek and M.~Kepski, ``Human fall detection on embedded platform using
  depth maps and wireless accelerometer,'' \emph{Computer Methods and Programs
  in Biomedicine}, 2014.

\bibitem{UP_Fall_2019}
L.~Martínez-Villaseñor, H.~Ponce, J.~Brieva, E.~Moya-Albor,
  J.~Núñez-Martínez, and C.~Peñafort-Asturiano, ``Up-fall detection
  dataset: A multimodal approach,'' \emph{Sensors}, 2019.

\bibitem{ESPINOSA2019103520}
R.~Espinosa, H.~Ponce, S.~Gutiérrez, L.~Martínez-Villaseñor, J.~Brieva, and
  E.~Moya-Albor, ``A vision-based approach for fall detection using multiple
  cameras and convolutional neural networks: A case study using the up-fall
  detection dataset,'' \emph{Computers in Biology and Medicine}, 2019.

\end{thebibliography}

\begin{IEEEbiography}[{\includegraphics[width=1in,height=1.25in,clip,keepaspectratio]{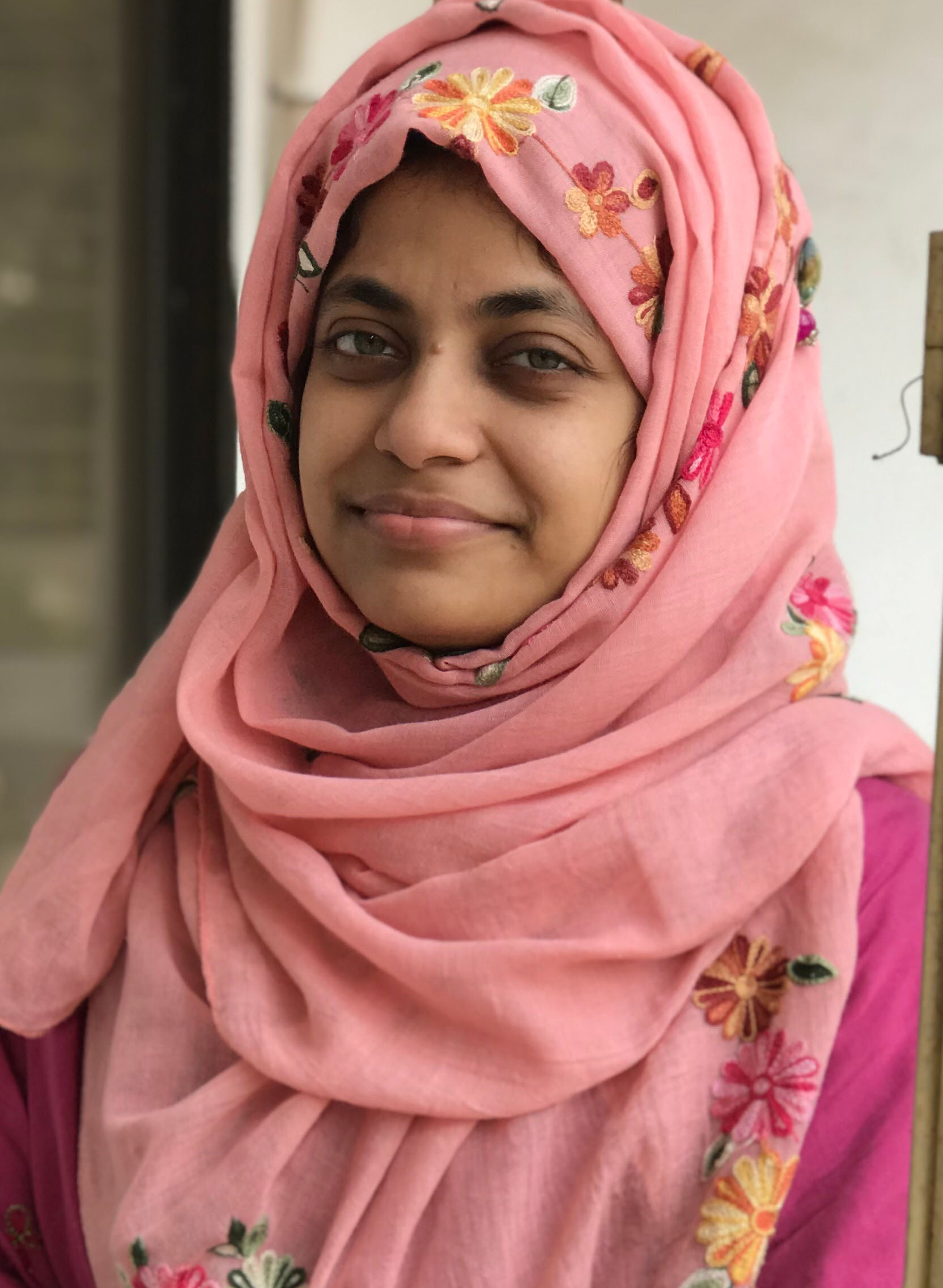}}]{Sania Zahan} received her B.Sc. in Engineering (Computer Science and Engineering) in 2015 from Pabna University of Science and Technology and her M.Sc. in Computer Science and Engineering in 2019 from Rajshahi University of Engineering and Technology, Bangladesh. She is pursuing her PhD degree in Computer Science at the department of Computer Science and Software Engineering at the University of Western Australia (UWA). Sania is the recipient of multiple merit scholarships and the Varendra University Center for Interdisciplinary Research (CIR) grant 2019-20. Her research interests include computer vision, human action analysis, Biomedical signal processing and deep learning. 
\end{IEEEbiography}

\begin{IEEEbiography}[{\includegraphics[width=1in,height=1.25in,clip,keepaspectratio]{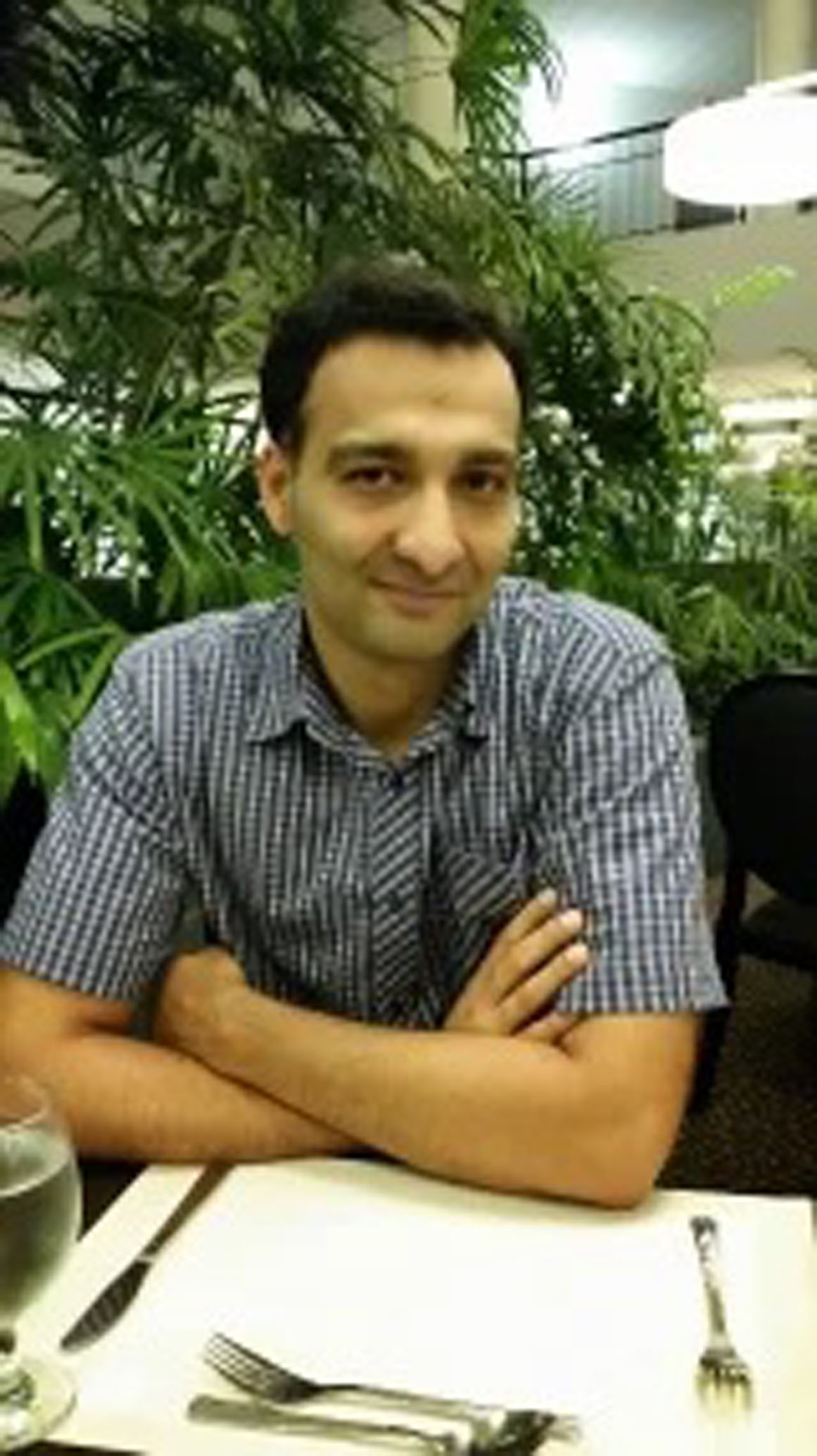}}]{Ghulum Mubashar Hassan} is a faculty member in the Department of Computer Science and Software Engineering at The University of Western Australia. Dr Hassan received his PhD in Computer Science \& Software Engineering and Civil \& Resource Engineering from UWA in 2016. He received his MS in in Electrical Engineering in 2004 from Oklahoma State University, USA and BS in Electrical Engineering from University of Engineering and Technology, Peshawar, Pakistan in 2001. His research interests are artificial intelligence, machine learning, and their applications in multidisciplinary problems. Dr Hassan is the recipient of multiple teaching excellence and research awards.
\end{IEEEbiography}

\begin{IEEEbiography}[{\includegraphics[width=1in,height=1.25in,clip,keepaspectratio]{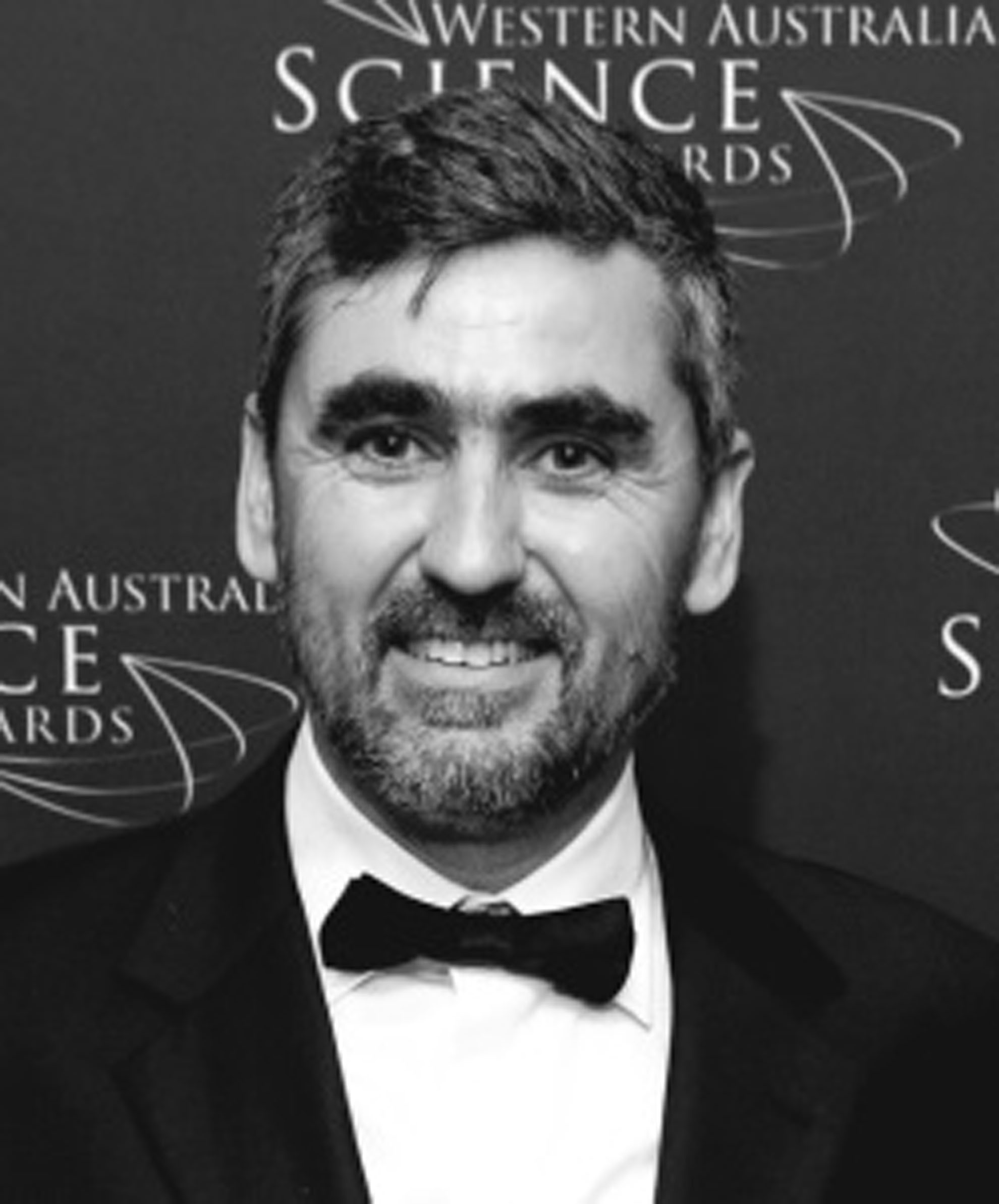}}]{Ajmal Mian} received his BB in Avionics from NED University Karachi in 1993, MS in Information Security from NUST in 2003 and PhD in Computer Science from UWA in 2007. Ajmal is a Professor of Computer Science at The University of Western Australia. He is the recipient of three prestigious national level fellowships from the Australian Research Council (ARC) including the Future Fellowship award. He is also a Fellow of the International Association for Pattern Recognition. He received the West Australian Early Career Scientist of the Year Award 2012, the HBF Mid-Career Scientist of the Year Award 2022 and several other awards including the Excellence in Research Supervision Award, EH Thompson Award, ASPIRE Professional Development Award, Vice-chancellors Mid-career Research Award, Outstanding Young Investigator Award, and the Australasian Distinguished Doctoral Dissertation Award. Ajmal Mian has secured research funding from the ARC, the National Health and Medical Research Council of Australia, US Department of Defence DARPA, and the Australian Department of Defence. He is a Senior Editor for IEEE Transactions on Neural Networks \& Learning Systems and an Associate Editor for IEEE Transactions on Image Processing and the Pattern Recognition journal. He served as a General Chair of the International Conference on Digital Image Computing Techniques \& Applications (DICTA 2019) and the Asian Conference on Computer Vision (ACCV 2018). His research areas include computer vision, deep learning, 3D point cloud analysis, facial recognition, human action recognition and video analysis.
\end{IEEEbiography}

\end{document}